\documentclass[runningheads]{llncs}
\usepackage[T1]{fontenc}
\usepackage[utf8]{inputenc}

\usepackage{graphicx}
\usepackage{booktabs}
\usepackage{multirow}
\usepackage{combelow}
\usepackage{tabularx}
\usepackage{xcolor}
\usepackage{lipsum}
\newcolumntype{Y}{>{\centering\arraybackslash}X}

\begin{document}    
\title{HistNERo: Historical Named Entity Recognition for the Romanian Language}
%
%

\author{Andrei-Marius Avram\textsuperscript{\rm 1},  Andreea Iuga\textsuperscript{\rm 1,*},
George-Vlad Manolache\textsuperscript{\rm 1,*},
Vlad-Cristian Matei\textsuperscript{\rm 1,*},
Răzvan-Gabriel Micliuş\textsuperscript{\rm 1,*},
Vlad-Andrei Muntean\textsuperscript{\rm 1,*},
Manuel-Petru Sorlescu\textsuperscript{\rm 1,*},
Dragoș-Andrei Șerban\textsuperscript{\rm 1,*}
Adrian-Dinu Urse\textsuperscript{\rm 1,*},
Vasile Păiș\textsuperscript{\rm 2},  and
Dumitru-Clementin Cercel\textsuperscript{\rm 1,a}
}

\def\thefootnote{*}\footnotetext{Authors are in alphabetical order. Equal contributions.}
\def\thefootnote{a}\footnotetext{Corresponding author.}
\def\thefootnote{\arabic{footnote}}

\authorrunning{A.-M. Avram et al.}


%

\institute{ \textsuperscript{\rm 1}Computer Science and Engineering Department, National University of Science and Technology POLITEHNICA Bucharest, Romania \\
\textsuperscript{\rm 2}Research Institute for Artificial Intelligence  “Mihai Drăgănescu”, Romanian Academy \\
\email{dumitru.cercel@upb.ro}
}

\maketitle              

\begin{abstract}

This work introduces HistNERo, the first Romanian corpus for Named Entity Recognition (NER) in historical newspapers. The dataset contains 323k tokens of text, covering more than half of the 19th century (i.e., 1817) until the late part of the 20th century (i.e., 1990). Eight native Romanian speakers annotated the dataset with five named entities. The samples belong to one of the following four historical regions of Romania, namely Bessarabia, Moldavia, Transylvania, and Wallachia. We employed this proposed dataset to perform several experiments for NER using Romanian pre-trained language models. Our results show that the best model achieved a strict F1-score of 55.69\%. Also, by reducing the discrepancies between regions through a novel domain adaption technique, we improved the performance on this corpus to a strict F1-score of 66.80\%, representing an absolute gain of more than 10\%.


\keywords{Named Entity Recognition  \and Historical Newspapers \and Romanian Language \and Novel Dataset \and Transformer.}
\end{abstract}

\section{Introduction}

A key area of artificial intelligence (AI) is natural language processing (NLP), concerned with how computers and human language interact. Its fundamental goal is to empower machines to comprehend, interpret, and reply to human languages in a valuable and meaningful manner \cite{avram2023multilingual}. This field has made significant contributions to the growth of AI by developing algorithms and models that facilitate machine understanding of text and speech, enabling applications such as automatic translation \cite{mohammadshahi2022small}, sentiment analysis \cite{echim2023adversarial}, and chatbots \cite{kim2024understanding}. 

Named Entity Recognition (NER) is a crucial part of nearly all NLP systems. It is responsible for identifying and categorizing entities, which include persons, organizations, dates, and organizations \cite{pakhale2023comprehensive}. In the context of historical documents, NER helps historians and researchers sift through vast amounts of textual data, automatically identifying and categorizing key entities. This speeds up the process of comprehending historical content and helps to extract insightful information and connections that could have been difficult or time-consuming to obtain through manual analysis alone \cite{ehrmann2022overview}.

As a testament to the growing importance of NER in historical research, we introduce a novel corpus - the \textbf{Hist}orical \textbf{N}amed \textbf{E}ntity \textbf{R}ecognition for the \textbf{Ro}manian language (HistNERo). This dataset addresses the unique linguistic nuances of the old Romanian language. It encompasses a diverse range of historical documents spanning various periods (i.e., from 1817 to 1990) from four regions where Romanian was historically spoken: Bessarabia, Moldavia, Transylvania, and Moldavia. HistNERo was meticulously annotated with five named entities (i.e., \texttt{PERSON}, \texttt{ORGANIZATION}, \texttt{LOCATION}, \texttt{PRODUCT}, and \texttt{DATE}), as proposed by the first edition of HIPE (Identifying Historical People, Places and other Entities) \cite{ehrmann2020overview}, employing a group of eight annotators, two for each region. The dataset comprises 10,026 sentences that contain 323,865 tokens and 9,601 annotated entities.
The dataset was gathered from the ROmanian DIachronic Corpus with Annotations (RODICA) corpus \cite{daniela2017recovering}.

Building upon the creation of the HistNERo dataset, we conducted extensive experiments to evaluate its effectiveness in conjunction with several Romanian large language models available in the literature. Our results outlined that the best-performing model is RoBERT-base \cite{masala2020robert}, which obtained a strict F1-score of 55.69\% on this dataset. Additionally, since the historical documents were collected from different historical regions of Romania, each one with its linguistical characteristics, we experimented with a novel domain adaptation technique called loss reversal, which reverses the sign of the domain discriminator loss instead of the gradient before entering the feature extractor, as proposed by Ganian et al. \cite{ganin2015unsupervised}. 
As the results suggest, employing this technique allows the large language models to better distinguish between the four regions. Thus, the performance of our best-performing model, RoBERT-base, was raised to 63.85\% strict F1-score and even surpassed by RoBERT-large, which obtained a strict F1-score of 66.80\%.

To summarize, our contribution is threefold:
\begin{itemize}
    \item We make available under an open license\footnote{\url{https://huggingface.co/datasets/avramandrei/histnero}} a historical corpus of 323k tokens annotated with five named entities by a group of annotators.
    \item We analyze the performance of various Romanian pre-trained language models on this dataset.
    \item We propose a new technique (called loss reversal) to improve the performance of most of the tested models by forcing them to learn embeddings that are more easily separable in the latent space.
\end{itemize}

\section{Related Work}
\label{sec:related}

\subsection{Named Entities in  Historical Documents}
Recent research on NER in historical texts highlighted the challenges posed by noisy and heterogeneous inputs \cite{10.1145/3604931}. As part of the Europeana Newspapers project, Neudecker \cite{neudecker-2016-open} detailed the creation of 100-page datasets for NER in historical newspapers for the languages Dutch, French, and German.
Hubková et al. \cite{hubkova-etal-2020-czech} presented a NER corpus of Czech historical newspapers. Acknowledging the problems associated with historical text, the authors propose an entity class named ambiguous for text spans; however, the annotator is unsure of the correct type.
Historical newspaper texts produced in French, German, Finnish, and Swedish between 1850 and 1950 are available in the NewsEye corpus \cite{10.1145/3404835.3463255}, which has been annotated for NER.

Additionally, to enable the performance comparison of NER on historical texts and to improve the robustness of current NER systems on non-standard inputs, the first edition of the HIPE competition (named HIPE-2020) \cite{ehrmann2020overview} was proposed in 2020. The participants had to forecast the same five entities as in HistNERo, and it was aimed at historical newspapers in French, German, and English. However, in addition to our task, the participants had to predict finer-grained entities, as well as link the entities to a unique referent in a knowledge database (i.e., Wikipedia or the NIL node if the referent for the mention was not located in the database). The La Rochelle University-affiliated L3i team generally obtained the highest scores, which approached the task in a multi-task setting using a hierarchical transformer-based model \cite{vaswani2017attention}.

The last HIPE competition (named HIPE-2022) \cite{ehrmann2022overview} was held in 2022, with the primary goal of providing new perspectives on the transferability of NER. The competition presented participants with the difficulties of multilingualism, as well as customizing their systems to domain-specific entities and various annotation sets. More specifically, HIPE-2022 focused on the same three tasks as HIPE-2020. In addition to working with historical newspapers, the  HIPE-2022 dataset also included classical commentaries; the former were gathered from five languages (English, German, French, Finnish, and Swedish), while the latter were collected from three languages (English, German, and French). However, compared to HIPE-2020, there were no clear winners on this dataset, with a different team ranking first on diverse subtasks.

\subsection{Romanian Named Entity Recognition}

Considering the Romanian language, ongoing activities aim to make historical documents available for research. In this context, the goal of the DeLORo project \cite{cristea2021} is to develop technology that can translate ancient printed and uncial Cyrillic-Romanian documents into Latin characters.

However, most NER research for the Romanian language is currently focused on contemporary language. RoNEC \cite{dumitrescu2020introducing} is a corpus covering standard NE classes and expressions (i.e., person, location, organization, and date) and additional NEs, such as money, quantity, and work of art. LegalNERo \cite{pais_vasile_2021_4922385} is a corpus with NEs in the Romanian legal domain, constructed primarily on legislation, extracted from the larger MARCELL-RO corpus \cite{varadi-etal-2020-marcell,tufis-etal-2020-collection}. In addition to common entity classes, it provides legal references. These were further refined by Costea et al. \cite{consilr2022legalnero}. NER systems in the legal domain were trained by \cite{pais-mitrofan-2021-towards-legalner,pais-etal-2021-named}. SiMoNERo \cite{mitrofan2019extragere,vergi5} is a NER extension to the MoNERo corpus \cite{mitrofan2019monero}. SiMoNERo contains NEs specific to the biomedical domain, such as anatomical parts, chemicals, disorders, and procedures. \cite{mitrofan-pais-2022-improving} proposed improving the recognition of biomedical NEs using a biologically inspired lateral inhibition approach. MicroBloggingNERo \cite{pais_vasile_2022_6905235,pais-EtAl:2022:CMLC10} is a corpus constructed based on social media text. It combines SiMoNERo and LegalNERo annotation schemes and adapts them to the social media context. A NER system based on MicroBloggingNERo was introduced by \cite{pais-etal-2022-romanian}.

\section{HistNERo Corpus}

\subsection{RELATE Platform}

Dataset annotation was realized in the RELATE platform\footnote{\url{https://relate.racai.ro}}. RELATE \cite{pais-etal-2020-processing} is a modern platform incorporating a large number of tools \cite{pais2020multiplepipelinesrelate} for processing the Romanian language. It was previously used for automatic annotation of large corpora, such as the MARCELL legislative corpus \cite{varadi-etal-2020-marcell,tufis-etal-2020-collection} and the CURLICAT corpus \cite{vradi-EtAl:2022:LREC}, as well as for creating Romanian language named entity corpora, such as MicroBloggingNERo and LegalNERo. The platform offers modules for corpus management, automatic annotations, and manual NE annotation. The manual NE annotation is accomplished at the span level employing the integrated BRAT component \cite{stenetorp-etal-2012-brat}. From the corpus management module, the annotator opens a document in the BRAT component and selects a text span with the mouse to which an NE class will be associated. The RELATE platform keeps track of the last annotated document, making it easier for the annotator to resume work or access the following document in his allocated batch.

\subsection{Annotation Process}

The annotation of the HistNERo dataset was performed by following the guidelines\footnote{\url{https://zenodo.org/records/3604227}}  of the HIPE-2020 competition, using a group of eight annotators, two for each region. When several entities were in a sequence, we annotated only the longest one. Also, we did not annotate nested entities. The annotation process took about four months and involved a process composed of four steps as follows:

\begin{enumerate}
\item In the first step of the process, the annotators of each region annotated the first two documents, resulting in an average Cohen's Kappa (CK) inter-annotator agreement (IAA) of 76.4\% and an average F1-score of 66.7\%, as depicted in Table \ref{tab1:ia_reg}. Also, as can be observed in Table \ref{tab:ia_ent}, the highest IAA was on the \texttt{DATE} entity, which usually is easier to identify and correctly annotate, even on historical documents. On the other hand, the \texttt{PRODUCT} entity is more challenging to determine in the old text due to the archaisms and regionalisms usually used to denote this category of entities. As a result, it obtained the lowest IAA score, i.e., a CK of 15.8\% and an F1-score of 7.3\%.

\item In the second step of the process, the annotators of each region discussed how they annotated the entities in the first two documents and then individually re-annotated the documents. This significantly boosted the IAA, resulting in an average 92.2\% CK and an average 86.3\% F1-score. Particularly, this discussion boosted the performance on the \texttt{PRODUCT} entity to a CK of 50.1\% and an F1-score of 38.6\%.

\item In the third step of the process, the rest of the documents in HistNERo were annotated, resulting in an average CK of 79.3\% and an average F1-score of 72.5\%.

\item In the fourth step of the process, the two annotators of each region agreed on the correct way of annotating the remaining entities in disagreement in HistNERo.
\end{enumerate}

\begin{table}[h]
\centering
\caption{For each of the four regions, the inter-annotator agreement on the first two documents before the first annotation discussion (i.e., Step 1), after the first annotation discussion (i.e., Step 2), and after all the documents have been annotated (i.e., Step 3).}
\begin{tabularx}{0.6\textwidth}{l| * {6}{Y}}
    \toprule
         \multirow{ 2}{*}{\textbf{Region}} & \multicolumn{2}{c|}{\textbf{Step 1}}  & \multicolumn{2}{c|}{\textbf{Step 2}}  & \multicolumn{2}{c}{\textbf{Step 3}} \\
         \cmidrule(lr){2-3} \cmidrule(l){4-5} \cmidrule(l){6-7}
         & \textbf{CK} & \textbf{F1} & \textbf{CK} & \textbf{F1} & \textbf{CK} & \textbf{F1} \\
         \midrule
         Bessarabia & 72.4 & 63.9 & 92.5 & 89.56 & 80.9 & 75.4\\
         Moldavia & 86.6 & 87.7 & 94.9 & 90.0 & 77.3 & 71.7\\
         Transylvania & 67.8 & 52.6 & 89.9 & 81.2 & 77.8 & 69.8\\
         Wallachia & 78.9 & 62.5 & 91.3 & 84.5 & 81.2 & 73.0\\
         \bottomrule
\end{tabularx}
\label{tab1:ia_reg}
\end{table}

\begin{table}[h]
\centering
\caption{For each of the five annotated named entities, the inter-annotator agreement on the first two documents before the first annotation discussion (i.e., Step 1), after the first annotation discussion (i.e., Step 2), and after all the documents have been annotated (i.e., Step 3).}
\begin{tabularx}{0.8\textwidth}{l| * {6}{Y}}
    \toprule
         \multirow{ 2}{*}{\textbf{Named Entity}} & \multicolumn{2}{c|}{\textbf{Step 1}}  & \multicolumn{2}{c|}{\textbf{Step 2}}  & \multicolumn{2}{c}{\textbf{Step 3}} \\
         \cmidrule(lr){2-3} \cmidrule(l){4-5} \cmidrule(l){6-7}
         & \textbf{CK} & \textbf{F1} & \textbf{CK} & \textbf{F1} & \textbf{CK} & \textbf{F1} \\
         \midrule
         PERSON & 71.2 & 65.4 & 95.6 & 90.4 & 82.1 & 77.3 \\
         ORGANISATION & 60.7 & 39.6 & 84.5 & 73.1 & 75.7 & 63.2 \\
         LOCATION & 75.2 & 83.5 & 96.0 & 90.9 & 80.8 & 83.0 \\
         PRODUCT & 15.8 & 7.3 & 50.1 & 38.6 & 44.0 & 27.6 \\
         DATE & 95.0 & 82.7 & 98.8 & 87.4 & 91.3 & 84.9 \\
         \bottomrule
\end{tabularx}
\label{tab:ia_ent}
\end{table}

\subsection{Data Preprocessing}

The output of the annotation processes consisted of files in the BRAT format that cannot be directly fed into a model; thus, they required preprocessing. The manner we approached this challenge was to tokenize the sentences in HistNERo using spaCy\footnote{\url{https://spacy.io}} and then automatically align the annotated entities based on their start and end character indexes. Moreover, we also annotate the resulting labels with the inside-outside-beginning format \cite{ramshaw1999text}, ensuring an unambiguous representation of the entities.

\subsection{HistNERo Statistics}

The resulting corpus contains 10,026 sentences and 323,865 tokens, annotated with 9,601 entities and 17,015 entity tokens, as depicted in Table \ref{tab:dataset_stats}. Most of the annotated entities belong to the \texttt{LOCATION} with 3,032 instances, closely followed by the \texttt{PERSON} entity with 2,914 instances, and the least amount of annotated entities belong to the \texttt{PRODUCT} entity with 424 instances. This distribution illustrates how most of the collected historical documents have focused more on people and places than on the goods those societies produced at the time.     

\begin{table}[h]
\centering
\caption{Named entity statistics on HistNERo.
}
\label{tab:dataset_stats}
\begin{tabularx}{0.82\textwidth}{l|c| * {3}{Y}}
    \toprule
         \multirow{ 2}{*}{\textbf{Named Entity}} & \multirow{ 2}{*}{\textbf{Region}} & \textbf{Total \# Tokens} & \textbf{Total \# 
         Entities} & \textbf{\# Tokens per Entity} \\
         \midrule
         \multirow{5}{*}{PERSON} & Bessarabia & 1,573 & 732 & 2.14 \\
          & Moldavia & 1,270 & 689 & 1.84 \\
          & Transylvania & 1,803 & 788 & 2.41 \\
          & Wallachia & 1,204 & 705 & 1.70 \\
          & \textbf{Total} & \textbf{5,950} & \textbf{2,914} & \textbf{2.04} \\
          \midrule
         \multirow{5}{*}{ORGANISATION} & Bessarabia & 1,268 & 481 & 2.63 \\
          & Moldavia & 253 & 145 & 1.74 \\
          & Transylvania & 1,048 & 509 & 2.05 \\
          & Wallachia & 1,768 & 1,283 & 1.37 \\
          & \textbf{Total} & \textbf{4,337} & \textbf{2,418} & \textbf{1.73} \\
          \midrule
         \multirow{5}{*}{LOCATION} & Bessarabia & 1,276 & 959 & 1.36 \\
          & Moldavia & 449 & 383 & 1.17 \\
          & Transylvania & 1,056 & 733 & 1.44 \\
          & Wallachia & 1,073 & 957 & 1.12 \\
          & \textbf{Total} & \textbf{3,854} & \textbf{3,032} & \textbf{1.27} \\
          \midrule
         \multirow{5}{*}{PRODUCT} & Bessarabia & 66 & 30 & 2.20 \\
          & Moldavia & 19 & 7 & 2.71 \\
          & Transylvania & 339 & 72 & 4.70 \\
          & Wallachia & 636 & 315 & 2.29 \\
          & \textbf{Total} & \textbf{1,060} & \textbf{424} & \textbf{2.50} \\
          \midrule
         \multirow{5}{*}{DATE} & Bessarabia & 456 & 186 & 2.45 \\
          & Moldavia & 319 & 151 & 2.11 \\
          & Transylvania & 574 & 250 & 2.29 \\
          & Wallachia & 465 & 226 & 2.05 \\
          & \textbf{Total} & \textbf{1,814} & \textbf{813} & \textbf{2.23} \\

          \midrule

          \textbf{Total} & - & \textbf{17,015} & \textbf{9,601} & \textbf{1.77} \\
           
    \bottomrule
\end{tabularx}
\end{table}

\subsection{Dataset Comparison}

Compared to other NER datasets available in the Romanian language (see Table \ref{tab:stats_comp}), HistNERO ranks second in terms of the number of annotated sentences and the number of total tokens, with 10,026 sentences and 323,865 tokens, being surpassed only by the second version of the Romanian Named Entity Corpus (RONECv2)\footnote{\url{https://github.com/dumitrescustefan/ronec}}, which has 12,330 sentences and 553,163 tokens. 
However, HistNERo includes the fewest annotated entities (i.e., 9,601). This number is influenced by both the frequency of entities in historical documents (HistNERo contains less than one entity per sentence) and by how many entities were used to annotate the corpus (e.g., 16 entities are used in RONECv1 and 15 in RONECv2).

\begin{table}[h]
\centering
\caption{HistNERo statistics (training, validation, test, and total) compared to other NER datasets in the Romanian language.}
\begin{tabularx}{0.98\textwidth}{l| * {5}{Y}}
    \toprule
         \multirow{ 2}{*}{\textbf{Dataset}} & \multirow{ 2}{*}{\textbf{\# Sents.}} & \multirow{ 2}{*}{\textbf{\# Tokens}} & \multirow{ 2}{*}{\textbf{\# Entities}} & \textbf{\# Entities Tokens} & \textbf{\# Tokens per Entity} \\
         \midrule
         RONECv1 \cite{dumitrescu2020introducing} & 5,127 & 180,467 & 26,377 & 50,760 & 2.13 \\
         RONECv2 & 12,330 & 553,163 & 80,283 & 50,760 & 2.13 \\
         LegalNERo \cite{puaislegalnero} & 8,284 & 264,335 & 13,614 & 54,456 & 4.00 \\
         MBNERo \cite{puaiș2022challenges} & 7,800 & 207,076 & 11,099 & 16,745 & 1.50 \\
         SiMoNERo \cite{mititelu2020romanian} & 5,418 & 163,707 & 15,493 & 25,563 & 1.65 \\
         \midrule
         \textbf{HistNERo-Train}  & \textbf{8,020} & \textbf{251,651} & \textbf{6,686} & \textbf{11,967} & \textbf{1.79} \\
         \textbf{HistNERo-Valid}  & \textbf{1,003} & \textbf{36,082} & \textbf{1,465} & \textbf{2,505} & \textbf{1.71} \\
         \textbf{HistNERo-Test}  & \textbf{1,003} & \textbf{36,132} & \textbf{1,485} & \textbf{2,543} & \textbf{1.71} \\
         \midrule
         \textbf{HistNERo-Total}  & \textbf{10,026} & \textbf{323,865} & \textbf{9,601} & \textbf{17,015} & \textbf{1.77} \\
         \bottomrule
\end{tabularx}
\label{tab:stats_comp}
\end{table}

\subsection{TF-IDF-based Data Analysis}

We further perform a Term Frequency-Inverse Document Frequency (TF-IDF) analysis on the documents of each region in HistNERo. The statistical technique known as TF-IDF consists of two parts: (1) TF measures the frequency with which a word appears in a document, and (2) IDF shows the term's commonality among all documents \cite{thapa2023nehate}. Thus, the TF-IDF score is the product of TF and IDF, with higher TF-IDF scores indicating words that are uncommon overall in the collection and frequent in a particular document.

We depict the results of this analysis in Table \ref{tab:tfidf} from which we can observe that several common words like "year(s)" or "government" obtain a high TF-IDF score in more than one region. Also, we can notice that the documents available for the Bessarabia region may contain a more nationalistic tone since the region's name or words like "country" are often used and are given higher importance by the TF-IDF algorithm.

\begin{table}[h]
\centering
\caption{TF-IDF analysis of the top five most common words found in each region of HistNERo.}
\begin{tabularx}{0.75\textwidth}{l|Y|Y|c}
    \toprule
         \textbf{Region} & \textbf{Romanian Words} & \textbf{Translation} & \hspace{0.7ex}\textbf{TF-IDF Score}\hspace{0.7ex} \\
         \midrule
         \multirow{5}{*}{\textbf{Bessarabia}} & basarabia & Bessarabia & 1.699 \\
          & ani & years & 1.138 \\
          & țării & country's & 1.086 \\
          & basarabiei & Bessarabia's & 1.081 \\
          & moldovei & Moldavia's & 1.061 \\
          \midrule
         \multirow{5}{*}{\textbf{Moldavia}} & lei & lions & 0.603 \\
          & guvernul & government & 0.476 \\
          & ziarul & newspaper & 0.476 \\
          & ani & year & 0.413 \\
          & aviatie & aviation & 0.407 \\
          \midrule
         \multirow{5}{*}{\textbf{Transylvania}}\hspace{0.7ex} & voru & want & 0.807 \\
          & politică & politics & 0.687 \\
          & anulu & year & 0.671 \\
          & romanu & Romanian & 0.663 \\
          & suntu & are & 0.660 \\
          \midrule
         \multirow{5}{*}{\textbf{Wallachia}} & făcut & done & 0.801 \\
          & românia & Romania & 0.697 \\
          & berlin & Berlin & 0.659 \\
          & guvernul & government & 0.633 \\
          & țara & country & 0.609 \\
    \bottomrule
\end{tabularx}
\label{tab:tfidf}
\end{table}

\section{Method}

\subsection{Baseline Models}

Regarding the architectures selected for training, transformer-based model fine-tuning has emerged as the de facto method for resolving NLP tasks in recent years \cite{bommasani2021opportunities}. In our methodology, we employ several such models trained specifically on Romanian text. 

Therefore, we assess four families of pre-trained language models for Romanian on HistNERo as follows:  the cased and uncased versions of BERT-ro (i.e., BERT-ro-c and BERT-ro-uc) \cite{dumitrescu2020birth}, the RoBERT versions (i.e., RoBERT-small, RoBERT-base, and RoBERT-large) \cite{masala2020robert}, the cased and uncased versions of the distilled Romanian BERT (i.e., DistRoBERT-c and DistRoBERT-uc) \cite{avram2022distilling}, and the three versions of the Romanian GPT (i.e., RoGPT-base, RoGPT-medium, and RoGPT-large) \cite{niculescu2021rogpt2}.

\subsection{Domain Adaptation}

Additionally, since the documents in HistNERo were collected from different historical regions of Romania, each with its linguistic features, we also employ a novel domain adaptation method that helps our foundational models learn region-independent features, boosting performance on the NER task.

We name this domain adaptation technique as \textit{loss reversal}, and it is inspired by the gradient reversal method in the domain adaptation technique \cite{ganin2015unsupervised}. More precisely, given a feature extractor $F$ whose role is to extract features from the input data, a label classifier $C$ whose role is to classify the features into the classes $y$, and a domain discriminator $D$ whose role is to classify the given features into the domains $d$, we reverse the gradient coming from the domain discriminator before backpropagating into the feature extractor. The following equations explain the gradient reversal \cite{ganin2015unsupervised}:

\begin{equation}
        \theta_C = \theta_C - \alpha\frac{\partial L_y}{\partial \theta_C}
\end{equation}

\begin{equation}
        \theta_{D} = \theta_{D} - \alpha\frac{\partial L_{d}}{\partial \theta_{D}}
\end{equation}

\begin{equation}
        \theta_F = \theta_F - \alpha \biggl(\frac{\partial L_y}{\partial \theta_F} - \lambda \frac{\partial L_{d}}{\partial \theta_F}\biggr)
\end{equation}
where $L_y$ is the label classifier's loss,  $\theta_C$ represents the label classifier's parameters, and $y$ are the labels. The domain discriminator's parameters are $\theta_{D}$, and the loss it incurs when predicting the domain labels $d$ is $L_{d}$. The feature extractor's parameters are $\theta_{F}$. Moreover, the learning rate is $\alpha$, and the hyperparameter $\lambda$ is used to scale the reversed gradients.

\begin{figure}[h!]
    \centering
    \includegraphics[width = 250pt, height = 200pt]{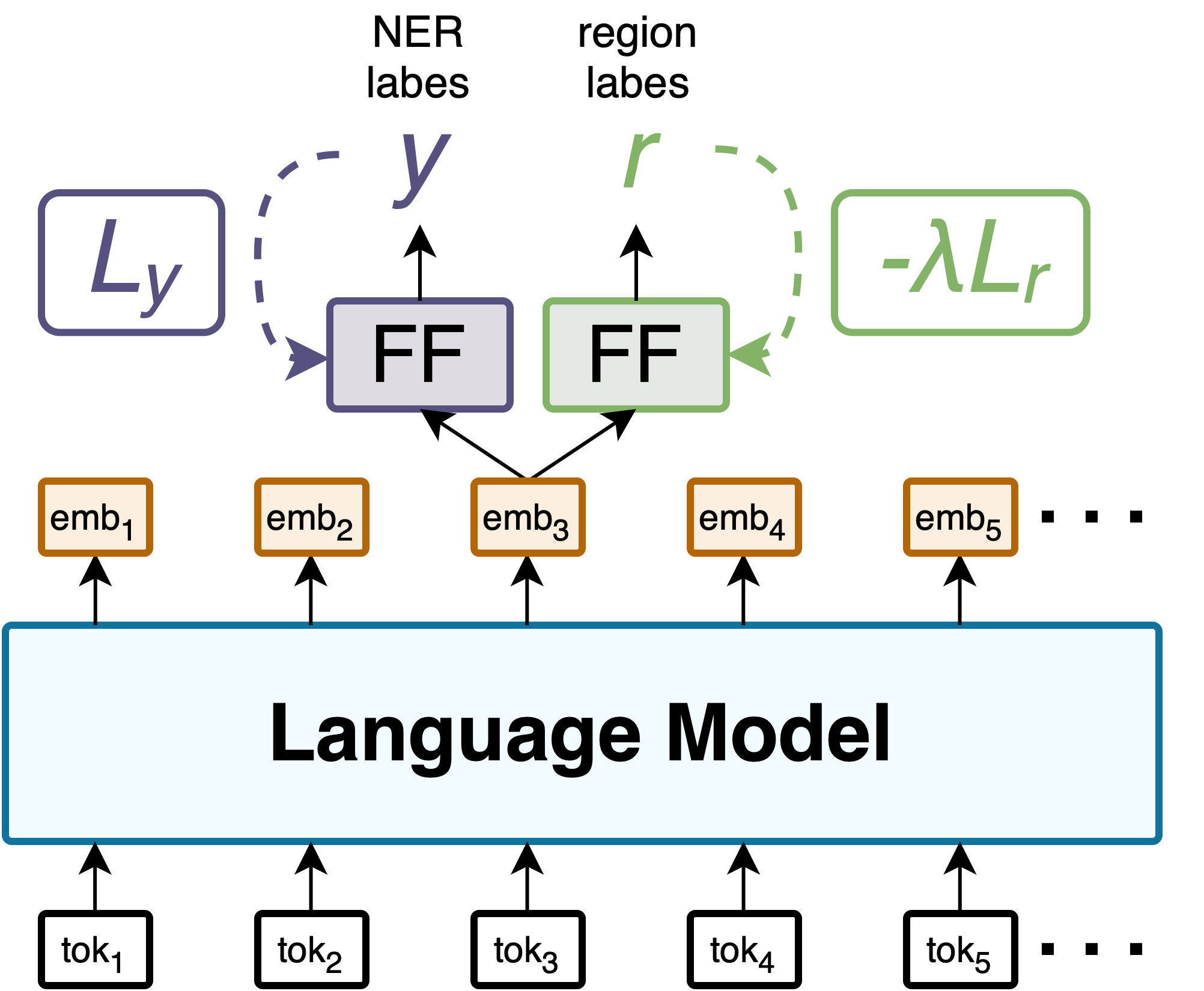}
    \caption{The loss reversal algorithm applied to the third token of a tokenized sentence. We note that FF stands for a feed-forward layer.}
    \label{fig:dom_adapt}
\end{figure}

To predict the NER label of each token using the domain adaption technique proposed in this paper, called loss reversal, we just switch the sign of the domain discriminator loss, as depicted in Figure \ref{fig:dom_adapt}.  In addition, the following equation provides a clear description of the loss reversal employed in this work:

\begin{equation}
    \mathcal{L}^i = \mathcal{L}_y^i - \lambda \mathcal{L}_r^i
\end{equation}
where $\mathcal{L}^i$ is the total loss of our system for the token $i$, $\mathcal{L}_y^i$ is the NER loss for the token $i$, $\mathcal{L}_r^i$ is the region loss for the token $i$, and $\lambda$ is a hyperparameter that controls the strength of the adversarial loss.

We believe that the loss reversal technique is much easier to implement than the gradient reversal in almost any framework, and, at least on the NER task applied to historical documents, we obtain better results, as it will be emphasized in Section \ref{sec:results}. However, more experiments on other tasks are required to confirm the general higher performance of our method, but such experiments would be beyond the scope of this paper.


\section{Implementation Details}

We train the models for 15 epochs, using the Adam optimizer \cite{KingBa15} with a learning rate of $2e-5$, a weight decay of $0.01$, and a batch size of 32 on an NVIDIA A100 GPU. Furthermore, we stabilize the training process by clipping \cite{pascanu2013difficulty} the gradient of the backpropagation algorithm to 2. We also employ the gradient checkpoint method to reduce memory consumption. To control the strength of the loss reversal, we set the hyperparameter $\lambda$ to 0.1, which we empirically observed to produce the best results in general. Furthermore, we split the HistNERo dataset into the train, validation, and test subsets, maintaining an 80\%, 10\%, and 10\% ratio. We made this dataset splitting publicly available to enhance reproducibility and the ease of comparison with our results.

\section{Results}
\label{sec:results}

The results of all the evaluated language models on HistNERo are depicted in Table \ref{tab:res_reg2} considering the four regions and the total performance. Also, Table 7 shows the performance of each of the five individual named entities. We measure the accuracy and the strict F1-score when we evaluate based on the regions and only the strict F1-score when we evaluate based on the named entities.

\begin{table}[ht!]
\centering
\label{tab:res_reg2}
\caption{The results of our evaluated language models on the four regions found in HistNERo: Bessarabia (Bess.), Moldavia (Mold.), Transylvania (Trans.), and Wallachia (Wall.), with the gradient reversal (Grad. Rev.) and loss reversal (Loss Rev.) domain adaptation algorithms.}
    \begin{tabularx}{\textwidth}{l| * {10}{Y}}
        \toprule
         \multirow{ 2}{*}{\textbf{Model}} & \multicolumn{2}{c|}{\textbf{Bess.}}  & \multicolumn{2}{c|}{\textbf{Mold.}}  & \multicolumn{2}{c|}{\textbf{Trans.}}  & \multicolumn{2}{c|}{\textbf{Wall.}} & \multicolumn{2}{c}{\textbf{Total}} \\
        \cmidrule(lr){2-3} \cmidrule(l){4-5} \cmidrule(l){6-7} \cmidrule(lr){8-9} \cmidrule(lr){10-11}
          & \textbf{Acc} & \textbf{F1} & \textbf{Acc} & \textbf{F1} & \textbf{Acc} & \textbf{F1} & \textbf{Acc} & \textbf{F1}  & \textbf{Acc} & \textbf{F1} \\
        \midrule
         BERT-ro-c \cite{dumitrescu2020birth}  & 97.60 &  66.93 &  99.30 & 60.46 & 94.69 & 39.77 & 97.07 & 53.11 & 97.18 & 55.03 \\
         \hspace{0.7ex}  w/ \textit{Grad. Rev.} & 97.92 &  74.60 & 97.98 & 52.27 & 95.74 & 51.08 & 97.67 & 65.04 & 97.35 & 62.51 \\
         \hspace{0.7ex}  w/ \textit{Loss Rev.} & \textbf{97.95} & 75.69 &  97.83 & 50.05 & 96.50 & \textbf{61.53} & 97.80 & 66.92 & 97.55 & 66.34 \\
         \midrule
         BERT-ro-uc \cite{dumitrescu2020birth}  & 97.42 & 63.24 & 99.10 & 40.09 & 94.22 & 31.95 & 96.87 & 52.96 & 96.93 & 50.57 \\
         \hspace{0.7ex}  w/ \textit{Grad. Rev.} & 97.06 & 66.91 &  97.13 & 30.58 & 96.41 & 49.35 & 97.27 & 60.71 & 96.77 & 56.61 \\
         \hspace{0.7ex}  w/ \textit{Loss Rev.} & 97.68 & 70.10 &  97.50 & 34.66 & 95.77 & 54.62 & 97.07 & 53.11 & 97.18 & 55.03 \\
        \midrule
         RoBERT-small \cite{masala2020robert}  & 97.57 &  66.93 &  98.68 & 48.14 & 93.96 & 41.62 & 97.02 & 50.37 & 96.87 & 53.03 \\
         \hspace{0.7ex}  w/ \textit{Grad. Rev.} & 97.03 & 66.81 &  97.90 & 3.73 & 95.29 & 42.10 & 97.39 & 55.06 & 96.92 & 53.80 \\
         \hspace{0.7ex}  w/ \textit{Loss Rev.} & 97.60 &  66.93 &  99.30 & 60.46 & 94.69 & 39.77 & 97.07 & 53.11 & 97.18 & 55.03 \\
         \midrule
         RoBERT-base \cite{masala2020robert} & 97.47 & 67.21 & \textbf{99.33} & \textbf{61.90} & 94.82 & 35.29 & 97.71 & 57.25 & 97.21 & 55.69 \\
         \hspace{0.7ex}  w/ \textit{Grad. Rev.} & 97.87 &  \textbf{78.74} &  97.98 & 42.50 & 96.62 & 57.00 & 97.69 & 65.04 & 97.56 & 64.97 \\
         \hspace{0.7ex}  w/ \textit{Loss Rev.} & 97.87 &  75.50 &  98.09 & 49.38 & 96.26 & 54.37 & 97.77 & 65.11 & 97.51 & 63.85 \\
         \midrule
         RoBERT-large \cite{masala2020robert} & 97.42 & 64.16 & 99.13 & 50.01 & 95.28 & 39.49 & 97.05 & 51.88 & 97.22 & 53.25 \\
         \hspace{0.7ex}  w/ \textit{Grad. Rev.} & 97.49 &  69.69 &  97.28 & 41.56 & 95.35 & 48.94 & 97.63 & 63.11 & 97.01 & 58.83 \\
         \hspace{0.7ex}  w/ \textit{Loss Rev.} & 98.00 &  76.67 &  97.83 & 45.45 & \textbf{96.68} & 59.72 & \textbf{97.90} & \textbf{69.96} & \textbf{97.64} & \textbf{66.80} \\
         \midrule
         DistRoBERT-c \cite{avram2022distilling}  & 96.28 & 55.97 & 97.79 & 44.18 & 95.80 & 53.11 & 97.37 & 52.30 & 96.81 & 52.86 \\
         \hspace{0.7ex}  w/ \textit{Grad. Rev.} & 94.98 & 46.54 &  97.57 & 48.57 & 95.32 & 43.90 & 96.36 & 38.23 & 96.00 & 42.73 \\
         \hspace{0.7ex}  w/ \textit{Loss Rev.} & 96.82 &  62.15 &  97.17 & 37.83 & 95.50 & 57.33 & 97.08 & 49.21 & 96.67 & 53.32 \\
         \midrule
         DistRoBERT-uc \cite{avram2022distilling} & 97.25 & 69.84 & 97.87 & 45.56 & 94.80 & 46.61 & 97.42 & 57.95 & 96.86 & 57.14 \\
         \hspace{0.7ex}  w/ \textit{Grad. Rev.} & 97.22 & 66.14 &  97.24 & 37.36 & 94.29 & 45.90 & 97.29 & 51.36 & 96.58 & 52.76 \\
         \hspace{0.7ex}  w/ \textit{Loss Rev.} & 97.33 &  69.67 &  97.68 & 44.15 & 94.56 & 46.53 & 97.10 & 52.34 & 96.69 & 54.99 \\
         \midrule
         RoGPT2-base \cite{niculescu2021rogpt2} & 96.76 & 62.81 & 97.35 & 35.41 & 95.32 & 40.00 & 97.63 & 54.61 & 96.83 & 51.15 \\
         \hspace{0.7ex}  w/ \textit{Grad. Rev.} & 96.33 & 55.36 &  97.39 & 38.63 & 94.98 & 38.84 & 97.27 & 50.38 & 96.54 & 47.66 \\
         \hspace{0.7ex}  w/ \textit{Loss Rev.} & 96.52 &  54.67 &  97.65 & 46.75 & 94.98 & 36.66 & 96.82 & 49.43 & 96.48 & 47.51 \\
         \midrule
         RoGPT2-med \cite{niculescu2021rogpt2} & 97.17 & 63.00 & 97.17 & 30.76 & 95.20 & 40.90 & 97.46 & 54.76 & 96.82 & 50.98 \\
         \hspace{0.7ex}  w/ \textit{Grad. Rev.} & 96.82 & 62.45 & 96.98 & 38.53 & 95.02 & 40.31 & 97.12 & 52.77 & 96.54 & 50.69 \\
         \hspace{0.7ex}  w/ \textit{Loss Rev.} & 96.28 &  57.24 &  97.39 & 46.31 & 95.17 & 38.59 & 97.18 & 49.62 & 96.53 & 48.78 \\
         \midrule
         RoGPT2-large \cite{niculescu2021rogpt2} & 97.09 & 66.66 & 97.32 & 36.17 & 94.89 & 40.32 & 97.63 & 56.71 & 96.81 & 51.01 \\
         \hspace{0.7ex}  w/ \textit{Grad. Rev.} & 96.55 &  58.62 &  97.35 & 41.58 & 94.92 & 40.57 & 97.11 & 51.18 & 96.49 & 49.57 \\
         \hspace{0.7ex}  w/ \textit{Loss Rev.} & 97.39 &  61.12 &  97.52 & 37.17 & 94.99 & 41.32 & 97.83 & 57.71 & 97.21 & 51.83 \\
        \bottomrule
    \end{tabularx}
\end{table}

\begin{table}[ht!]
\centering
\label{tab:resennt2}
\caption{The strict F1-scores of our evaluated language models on the five named entities (i.e., \texttt{PERS} → \texttt{PERSON}, \texttt{ORG} → \texttt{ORGANISATION}, \texttt{LOC} → \texttt{LOCATION}, \texttt{PROD} → \texttt{PRODUCT}, and \texttt{DATE}) found in HistNERo, with the gradient reversal (Grad. Rev.) and loss reversal (Loss Rev.) domain adaptation algorithms.}
    \begin{tabularx}{0.8\textwidth}{l| * {5}{Y}}
        \toprule
         \textbf{Model} & \textbf{PERS}  & \textbf{ORG}  & \textbf{LOC}  & \textbf{PROD}  & \textbf{DATE} \\
         \midrule
         BERT-ro-c \cite{dumitrescu2020birth} & 54.76 & 53.68 & 65.83 & 21.05 & 40.57 \\
         \hspace{0.7ex}  w/ \textit{Grad. Rev.} & 62.29 & 55.14 & 69.70 & 40.13 & 75.40 \\
         \hspace{0.7ex}  w/ \textit{Loss Rev.} & 66.66 & \textbf{61.12} & 71.42 & 41.17 & 78.12 \\
         \midrule
         BERT-ro-uc \cite{dumitrescu2020birth} & 40.96 & 52.63 & 66.10 & 10.81 & 35.29 \\
         \hspace{0.7ex}  w/ \textit{Grad. Rev.} & 55.31 & 51.13 & 61.53 & 32.25 & 72.72 \\
         \hspace{0.7ex}  w/ \textit{Loss Rev.} & 54.76 & 53.68 & 65.83 & 21.05 & 40.57 \\
         \midrule
         RoBERT-small \cite{masala2020robert} & 47.05 & 45.74 & 69.44 & 24.00 & 41.93 \\
         \hspace{0.7ex}  w/ \textit{Grad. Rev.} & 51.79 & 45.83 & 64.62 & 26.66 & 65.62 \\
         \hspace{0.7ex}  w/ \textit{Loss Rev.} & 54.76 & 53.68 & 65.83 & 21.05 & 40.57 \\
         \midrule
         RoBERT-base \cite{masala2020robert} & 58.22 & 56.15 & 66.93 & 10.85 & 32.23 \\
         \hspace{0.7ex}  w/ \textit{Grad. Rev.} & \textbf{68.31} & 53.08 & 70.99 & \textbf{46.15} & 81.25 \\
         \hspace{0.7ex}  w/ \textit{Loss Rev.} & 64.46 & 56.30 & 69.29 & 44.44 & \textbf{81.96} \\
         \midrule
         RoBERT-large \cite{masala2020robert} & 55.71 & 47.31 & 63.56 & 10.81 & 41.26 \\
         \hspace{0.7ex}  w/ \textit{Grad. Rev.} & 58.48 & 49.79 & 69.78 & 33.35 & 73.01 \\
         \hspace{0.7ex}  w/ \textit{Loss Rev.} & 67.43 & 60.60 & \textbf{74.67} & 32.43 & 76.19 \\
         \midrule
         DistRoBERT-c \cite{avram2022distilling} & 52.41 & 44.94 & 61.66 & 30.30 & 65.67 \\
         \hspace{0.7ex}  w/ \textit{Grad. Rev.} & 38.09 & 30.43 & 54.48 & 12.15 & 52.16 \\
         \hspace{0.7ex}  w/ \textit{Loss Rev.} & 58.13 & 39.63 & 66.13& 22.05 & 73.08 \\
         \midrule
         DistRoBERT-uc \cite{avram2022distilling} & 58.92 & 47.30 & 64.73 & 27.50 & 73.33 \\
         \hspace{0.7ex}  w/ \textit{Grad. Rev.} & 49.63 & 56.78 & 66.47 & 16.39 & 63.76 \\
         \hspace{0.7ex}  w/ \textit{Loss Rev.} & 50.57 & 47.74 & 66.93 & 17.76 & 70.96 \\
         \midrule
         RoGPT2-base \cite{niculescu2021rogpt2} & 44.36 & 47.35 & 64.82 & 27.58 & 52.45 \\
         \hspace{0.7ex}  w/ \textit{Grad. Rev.} & 37.53 & 44.25 & 62.50 & 16.21 & 59.74 \\
         \hspace{0.7ex}  w/ \textit{Loss Rev.} & 41.73 & 41.76 & 62.55 & 13.33 & 51.42 \\
         \midrule
         RoGPT2-med \cite{niculescu2021rogpt2} & 48.05 & 39.65 & 65.58 & 16.66 & 64.61 \\
         \hspace{0.7ex}  w/ \textit{Grad. Rev.} & 46.10 & 42.90 & 67.20 & 27.90 & 55.96 \\
         \hspace{0.7ex}  w/ \textit{Loss Rev.} & 44.03 & 41.89 & 63.77 & 10.05 & 50.13 \\
         \midrule
         RoGPT2-large \cite{niculescu2021rogpt2} & 47.26 & 40.81 & 66.14 & 29.41 & 58.46 \\
         \hspace{0.7ex}  w/ \textit{Grad. Rev.} & 46.70 & 51.37 & 64.75 & 20.00 & 56.75 \\
         \hspace{0.7ex}  w/ \textit{Loss Rev.} & 48.27 & 43.14 & 66.30 & 12.5 & 57.14 \\
        \bottomrule
    \end{tabularx}
\end{table}

The highest overall accuracy was obtained by RoBERT-large with 97.22\%, and the highest overall strict F1-score was obtained by DistilRoBERT-uc with 57.14\%. When it comes to each individual region, the highest strict F1-score was obtained by DistilRoBERT-uc for the Bessarabia region with 69.84\%, by RoBERT-base for the Moldavia region with 61.90\%, by DistilRoBERT-c for the Transylvania region with 53.11\%, and by DistilRoBERT-uc for the Wallachia region with 57.95\%. On the other hand, when considering the performance on each individual category of named entities, the highest strict F1-score was obtained by DistRoBErT-uc for the \texttt{PERSON} entity with 58.92\%, by RoBERT-base for the \texttt{ORGANIZATION} entity with 56.15\%, by RoBERT-small for the \texttt{LOCATION} entity with 69.44\%, by DistilRoBERT-c for the \texttt{PRODUCT} entity with 30.30\%, and by DistRoBERT-uc for the \texttt{DATE} entity with 73.33\%.

By incorporating the two domain adaptation techniques we experimented with in this work, the gradient and loss reversal, we significantly boosted the performance of our language models. Specifically, we improved the performance of 7 out of 10 models, with an average relative improvement of 1.40\% for the gradient reversal and an average relative improvement of 6.36\% for the loss reversal. Furthermore, we achieved a new highest overall accuracy of 97.64\% and a new highest overall strict F1-score of 66.80\% with RoBERT-large and the loss reversal algorithm. When considering each region, a significant boost can be observed in the Transylvania region, which obtained a relative strict F1-score score improvement of 16.34\% with the gradient reversal algorithm and a relative strict F1-score improvement of 20.63\% with the loss reversal algorithm. On the other hand, the lowest relative strict F1-scores improvements were observed for the Moldavia region with -12.63\% when the gradient reversal was employed and 2.32\% when the loss reversal was used.

By looking at the relative strict F1-scores of each named entity, we can observe that the highest average improvements were obtained on the \texttt{DATE} named entity with 72.79\% for the gradient reversal algorithm and 47.61\% for the loss reversal algorithm. We believe that this is because this entity is the easiest to standardize between regions compared to other more region-specific named entities such as \texttt{PERSON} or \texttt{ORGANIZATION}, and it benefits much more from domain adversarial training whose aim in this work is to create region independent features. On the other hand, the lowest strict F1-scores improvements were obtained on the \texttt{PRODUCT} named entity with -0.43\% for the gradient reversal algorithm and 2.85\% for the loss reversal algorithm. We believe that low improvement is due to the rarity of this named entity in the dataset, which affects the overall performance and possible refinements. Other strategies for class imbalance should be used to significantly improve the performance on the \texttt{PRODUCT} named entity.

\subsection{Inter-Regional Evaluation}

\begin{figure}[ht!]
    \centering
    \includegraphics[width=0.85\textwidth]{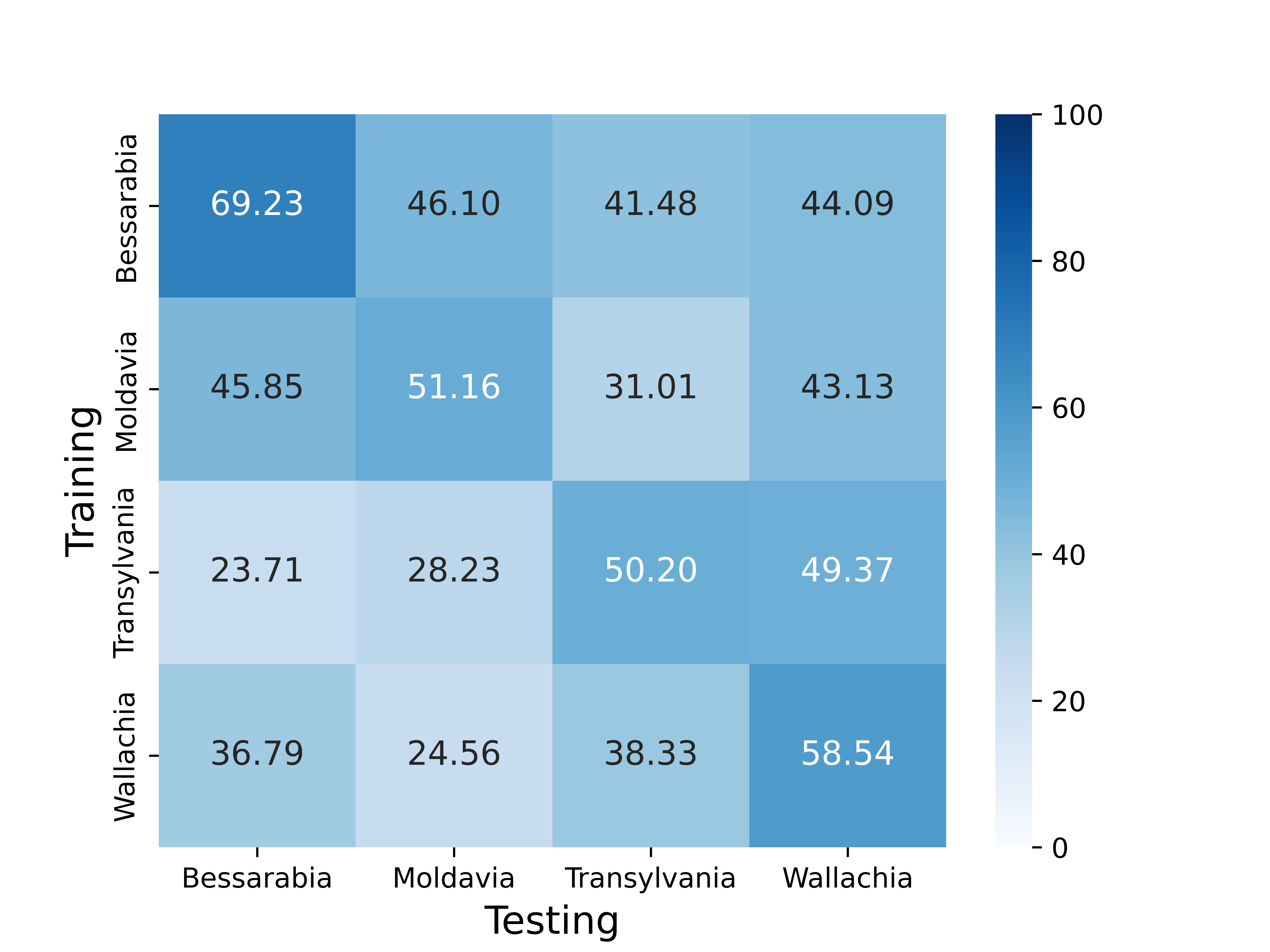}
    \caption{Inter-regional strict F1-scores of the RoBERT-large model.}
    \label{fig:inter_reg_eval}
\end{figure}

We also tested the inter-regional performance of RoBERT-large (i.e., the best-performing model in Table \ref{tab:res_reg2}) by training it on the subset determined by each region and then evaluating the strict F1-score of the model on each of the four areas. The results are depicted in Figure \ref{fig:inter_reg_eval}. We can observe that the highest strict F1-scores were obtained on the main diagonal, which is a regional evaluation of the model (i.e., trained and evaluated on the same region), with the Bessarabia region leading the benchmark and obtaining a strict F1-score of 69.23\%. On the other hand, the highest inter-regional strict F1-score was obtained when RoBERT-large was trained on the Transylvania region and evaluated on the Wallachia region, achieving a strict F1-score of 49.37\%, at less than 1\% difference compared to the intra-regional result, showing the possible similarities between the two geographical areas. Similarly, we can also observe that the results of the inter-regional evaluation between the Bessarabia and Moldavia regions are also well correlated, demonstrating once more the historical closer ties that existed between the two areas compared to other parts of Romania.

\subsection{T-SNE Visualizations}

\begin{figure}[ht!]
    \centering
    \includegraphics[width=0.49\textwidth]{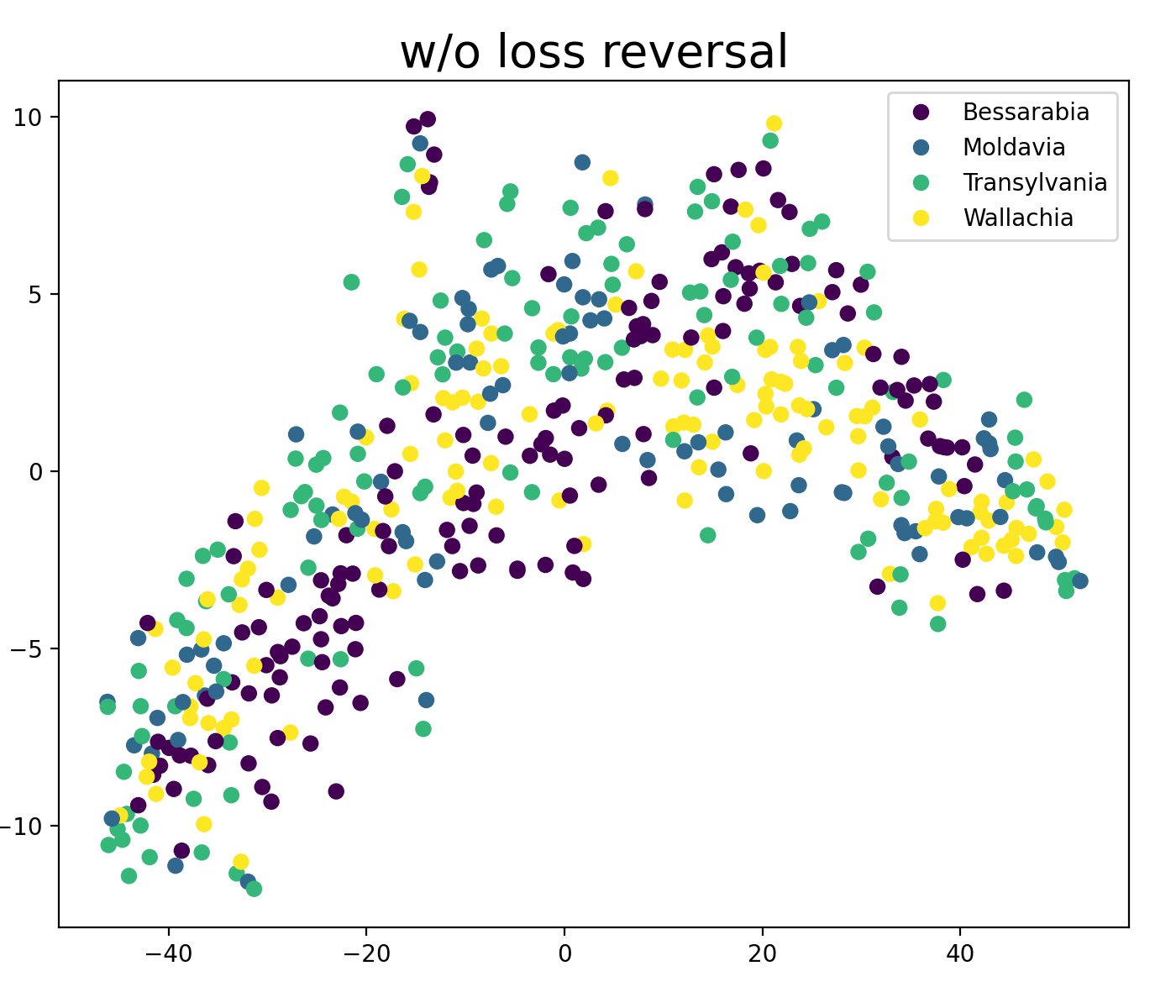}
    \includegraphics[width=0.49\textwidth]{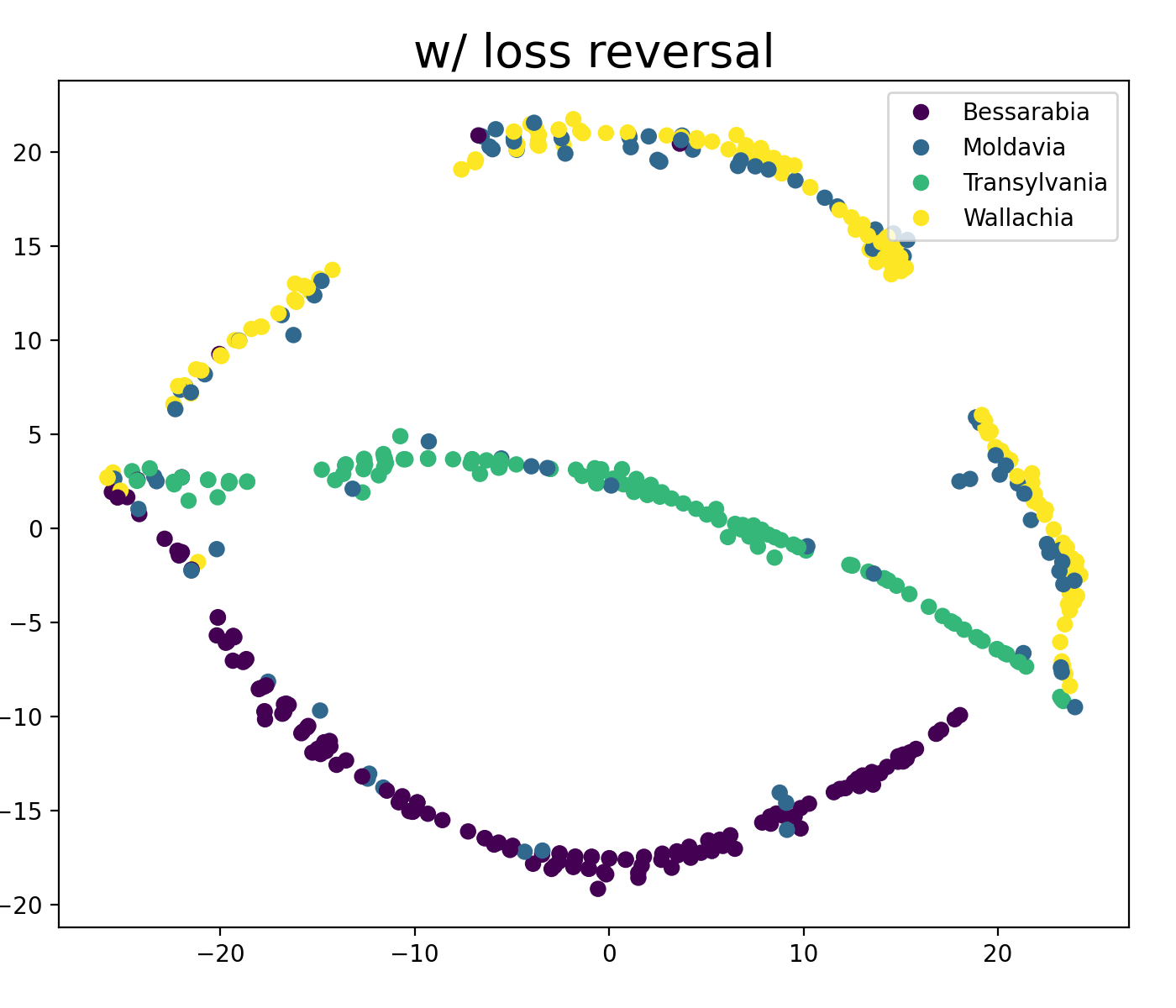}
    \caption{Clustering of the HistNERo test set embeddings produced by the BERT-large model using t-SNE with and without loss reversal.}
    \label{fig:enter-label}
\end{figure}

We further analyze the embeddings produced by BERT-large on the HistNERo test set to understand the effects of the loss reversal algorithm. Thus, we average the embeddings produced by BERT-large on each token and visualize the results using t-SNE \cite{van2008visualizing}. The results are depicted in Figure \ref{fig:enter-label}. We can notice that without the loss reversal, there is no clear separation between the embeddings, making the model less robust when challenged with linguistic variations from the Romanian regions. On the other hand, when the loss reversal algorithm is employed, we can easily observe a clear separation in the embedding space between the Bessarabia, Transylvania, and Wallachia regions. However, BERT-large struggles to create a separate cluster for the Moldavia region. We believe that this is because Moldavia and Bessarabia were part of the same country for much of their history, and there are few linguistic differences between them. Thus, the model struggled to create another cluster in the embedding space for the Moldavia region.

\section{Conclusion and Future Work}

Analyzing historical documents is essential in research, providing vital insights for a nuanced understanding of the past. Thus, we introduced HistNERo in this work, a valuable resource that will advance the state of the art in historical NER, facilitating a more accurate and nuanced analysis of Romanian historical documents and contributing to a deeper understanding of the country's historical evolution. Additionally, since the HistNERo corpus contains documents collected from various historical regions of Romania, we developed a novel domain adaptation that adds the region prediction loss to the overall optimization target but with the reversed sign. Our experimental results outlined that, by incorporating this simple technique, the performance of our models relatively improved by an average of 6.36\%, the best model with the loss reversal achieved a strict F1-score of 66.80\%, an absolute improvement of more than 10\% compared to the best model that did not employ any domain adaptation technique. Finally, we intend to incorporate both the HistNERo dataset and the results obtained by the evaluated models in the LiRo benchmark \cite{dumitrescu2021liro}.

\section*{Acknowledgements}
This work was supported by a grant from the National Program for Research of the National Association of Technical Universities - GNAC ARUT 2023.

%
%
%
\bibliographystyle{splncs04}
\bibliography{mybibliography}
\end{document}